\pdfoutput=1
\documentclass[11pt]{article}

\usepackage{acl}

\usepackage{comment}
\usepackage{times}
\usepackage{siunitx}
\usepackage{latexsym}
\usepackage{multirow}
\usepackage{graphicx}
\usepackage{amsmath,amssymb}
\usepackage{booktabs}
\usepackage{enumitem} 
\usepackage{txfonts}
\usepackage{booktabs}
\usepackage[T1]{fontenc}
\usepackage[utf8]{inputenc}

\usepackage{microtype}
\usepackage{xcolor,colortbl}

\definecolor{best}{HTML}{ccffe6}
\definecolor{worst}{HTML}{ffd1b3}

\DeclareMathOperator{\EX}{\mathbb{E}}% expected value
\title{FairLex: A Multilingual Benchmark for Evaluating \\ Fairness in Legal Text Processing}

\author{Ilias Chalkidis$^{\;\dagger}$\thanks{\;\;Corresponding author: \texttt{ilias.chalkidis@di.ku.dk}}\qquad Tommaso Pasini$^{\;\dagger}$ \qquad Sheng Zhang$^{\;\diamond}$ \\
\textbf{Letizia Tomada}$^{\;\ddagger}$\qquad \textbf{Sebastian Felix Schwemer}$^{\;\ddagger}$ \qquad \textbf{Anders Søgaard}$^{\;\dagger}$ \\
$^{\dagger\;}$ Department of Computer Science, University of Copenhagen, Denmark \\ $^{\ddagger\;}$Faculty of Law, University of Copenhagen, Denmark\\ $^{\diamond\;}$National University of Defense Technology,  People's Republic of China \\}

\begin{document}

\maketitle

\begin{abstract}
We present a benchmark suite of four datasets for evaluating the fairness of pre-trained language models and the techniques used to fine-tune them for downstream tasks. Our benchmarks cover four jurisdictions (European Council, USA, Switzerland, and China), five languages (English, German, French, Italian and Chinese) and fairness across five attributes (gender, age, region, language, and legal area). In our experiments, we evaluate pre-trained language models using several group-robust fine-tuning techniques and show that performance group disparities are vibrant in many cases, while none of these techniques guarantee fairness, nor consistently mitigate group disparities. Furthermore, we provide a quantitative and qualitative analysis of our results, highlighting open challenges in the development of robustness methods in legal NLP. 
\end{abstract}

\section{Introduction}
\label{sec:introduction}

Natural Language Processing (NLP) for law \cite{chalkidis-kampas-2019-dlaw,aletras-etal-2019-nllp, zhong-etal-2020-nlp,chalkidis-2022-lexglue} receives increasing attention. 
Assistive technologies can speed up legal research or discovery significantly assisting lawyers, judges and clerks. They can also help legal scholars to study case law \cite{Katz2012,coupette2021}, improve access of law to laypersons, help sociologists and research ethicists to expose biases in the justice system \cite{angwin2016,dressel2018}, and even scrutinize decision-making itself \cite{recon}.

In the context of law, the principle of \emph{equality} and \emph{non-discrimination} is of paramount importance, although its definition varies at international, regional and domestic level. For example, EU non-discrimination law prohibits both direct and indirect discrimination. Direct discrimination occurs when one person is treated \emph{less favourably than others would be treated in comparable situations} on grounds of sex, racial or ethnic origin, disability, sexual orientation, religion or belief and age.\footnote{An in-depth analysis of the notion of discrimination and fairness in law is presented in Appendix~\ref{sec:appendix_discrimination}.} Given the gravity that legal outcomes have for individuals, assistive technologies cannot be adopted to speed up legal research at the expense of fairness \cite{Wachter2021BiasPI}, potentially also decreasing the trust in our legal systems  \cite{barfield2020}.

\begin{figure}[t]
    \centering
    \resizebox{0.95\columnwidth}{!}{
    \includegraphics{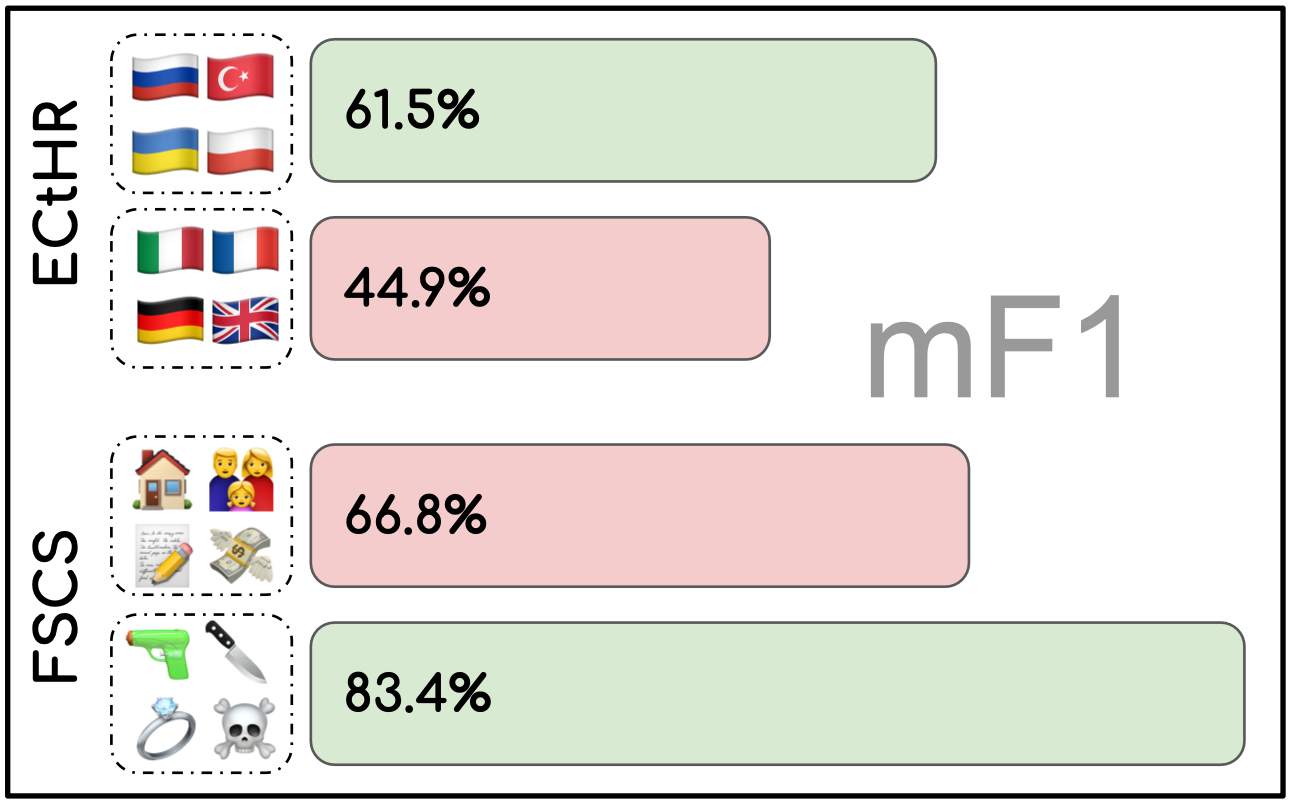}
    }
    \vspace{-2mm}
    \caption{\emph{Group disparity} for \emph{defendant state} (C.E. Europe vs. The Rest) in ECtHR and \emph{legal area} (Civil law vs. Penal law) in FSCS.}
    \label{fig:logo}
    \vspace{-6mm}
\end{figure}

Societal transformations perpetually shape our legal systems. The topic deserves great attention because AI systems learning from historical data pose the risk of lack of generalisability beyond the training data, and more importantly transporting biases previously encumbered in the data in future decision-making, thereby exponentially increasing their effect \cite{Delacroix_2022}. 

Historical legal data do not represent all groups in our societies equally and tend to reflect social biases in our societies and legal institutions. When models are deployed in production, they may reinforce these biases. For example, criminal justice is already often strongly influenced by racial bias, with people of colour being more likely to be arrested and receive higher punishments than others, both in the USA\footnote{\url{https://tinyurl.com/4cse552t}} and in the  UK.\footnote{\url{https://tinyurl.com/hkff3zcb}}

In recent years, the NLP and machine learning literature has introduced fairness objectives, typically derived from the Rawlsian notion of \emph{equal opportunities} \cite{rawls_theory_1971}, to evaluate the extent to which models discriminate across protected attributes.
Some of these rely on notions of resource allocation, i.e., reflecting the idea that groups are treated fairly if they are equally represented in the training data used to induce our models, or if the same number of training iterations is performed per group. This is sometimes referred to as the {\em resource allocation}~perspective on fairness \cite{lundgard-20}. Contrary, there is also a {\em capability}-centered approach to fairness \cite{anderson99point,robeyns2008justice}, in which the goal is to reserve enough resources per group to achieve similar performance levels, which is ultimately what is important for how individuals are treated in legal processes. We adopt a capability-centered approach to fairness and define fairness in terms of \emph{performance parity} \cite{pmlr-v80-hashimoto18a} or \emph{equal risk} \cite{donini18empirical}.\footnote{The dominant alternative to equal risk is to define fairness in terms of \emph{equal odds}. Equal odds fairness does not guarantee Rawlsian fairness, and often conflicts with the rule of law.}

Performance disparity \cite{pmlr-v80-hashimoto18a} refers to the phenomenon of high overall performance, but low performance on minority groups, as a result of minimizing risk across samples (not groups). Since some groups benefit more than others from models and technologies that exhibit performance disparity, this likely widens gaps between those groups. Performance disparity works against the ideal of fair and equal opportunities for all groups in our societies. We therefore define a {\em fair} classifier as one that has similar performance (equal risk) across all groups  \cite{donini18empirical}.

In sum, we adopt the view that (approximate) equality under the law in a modern world requires that our NLP technologies exhibit (approximately) equal risk across sensitive attributes. For everyone to be treated equally under the law, regardless of race, gender, nationality, or other characteristics, NLP assistive technologies need to be (approximately) insensitive to these attributes.
We consider three types of attributes in this work:

\begin{itemize}[leftmargin=8pt,itemsep=0.2mm,parsep=0.4mm]
    \item \emph{Demographics}:  The first category includes demographic information of the involved parties, e.g., the gender, sexual orientation, nationality, age, or race of the plaintiff/defendant in a case. In this case, we aim to mitigate biases against specific groups, e.g., a model performs worse for female defendants or is biased against black defendants. We can further consider information involving the legal status of involved parties, e.g., person vs. company, or private vs. public.
    \item \emph{Regional}:  The second category includes regional information, for example the courts in charge of a case. In this case, we aim to mitigate disparity in-between different regions in a given jurisdiction, e.g., a model performs better in specific cases originated or ruled in courts of specific regions. 
    \item  \emph{Legal Topic}: The third category includes legal topic information on the subject matter of the controversy. In this case, we aim to mitigate disparity in-between different topics (areas) of law, e.g., a model performs better in a specific field of law, for example penal cases.
\end{itemize}

\paragraph{Contributions}
We introduce FairLex, a multilingual fairness benchmark of four legal datasets covering four jurisdictions (Council of Europe, United States of America, Swiss Confederation and People's Republic of China), five languages (English, German, French, Italian and Chinese) and various sensitive attributes (gender, age, region, etc.). 
We release four pre-trained transformer-based language models, each tailored for a specific dataset (task) within our benchmark, which can be used as baseline models (text encoders). 
We conduct experiments with several group-robust algorithms and provide a quantitative and qualitative analysis of our results, highlighting open challenges in the development of robustness methods in legal NLP. 

\section{Related Work}

\paragraph{Fair machine learning} The literature on inducing approximately fair models from biased data is rapidly growing. See \newcite{10.1145/3457607,maklouf2021,ding2021retiring} for recent surveys. We rely on this literature in how we define fairness, and for the algorithms that we compare in our experiments below. As already discussed, we adopt a capability-centered approach to fairness and define fairness in terms of performance parity \cite{pmlr-v80-hashimoto18a} or equal risk \cite{donini18empirical}. The fairness-promoting learning algorithms we evaluate are discussed in detail in Section~\ref{sec:algorithms}. Some of these -- Group Distributionally Robust Optimization \cite{sagawa-etal-2020-dro} and Invariant Risk Minimization \cite{arjovsky-etal-2020-irm} -- have previously been evaluated for fairness in the context of hate speech \cite{wilds2021}. 

\paragraph{Fairness in law}

Studying fair machine learning in the context of legal (computational) applications has a limited history. In a classic study, \citet{angwin2016} analyzed the performance of the Correctional Offender Management Profiling for Alternative Sanctions (COMPAS) system, which was used for parole risk assessment (recidivism prediction) in the US. The system relied on 137 features from questionnaires and criminal records. \citeauthor{angwin2016} found that blacks were almost twice as likely as whites to be mislabeled as high risk (of re-offending), revealing a severe racial bias in the system.
The system was later compared to crowd-workers in \citet{dressel2018}. 
These studies relied on tabular data and did not involve text processing (e.g., encoding case facts or decisions).

More recently, \citet{wang-etal-2021-equality} studied legal judgment consistency using a dataset of Chinese criminal cases. They evaluated the consistency of LSTM-based models across region and gender and reported severe fairness gaps across gender. They also found that the fairness gap was particular severe for more serious crimes.
Another line of work \cite{rice2019,baker-gillis-2021-sexism,gumusel2022} explores representational bias with respect to race and gender analyzing word latent representations trained in legal text corpora. While we agree that representational bias can potentially reinforce unfortunate biases, these may not impact the treatment of individuals (or groups). We therefore focus on directly measuring equal risk on downstream applications instead.

Previous work has focused on the analysis of specific cases, languages or algorithms, but FairLex aims at easing the development and testing of bias-mitigation models or algorithms within the legal domain. FairLex allows researchers to explore fairness across four datasets covering four jurisdictions (Council of Europe, United States of America, Swiss Confederation and People's Republic of China), five languages (English, German, French, Italian and Chinese) and various sensitive attributes (gender, age, region, etc.). Furthermore, we provide competitive baselines including pre-trained transformer-based language models, adapted to the examined datasets, and an in-dept examination of performance of four group robust algorithms described in detail in Section~\ref{sec:algorithms}.

\begin{table*}[t]
    \centering
    \resizebox{\textwidth}{!}{
    \begin{tabular}{l|l|l|c|l|c}
    \multirow{2}{*}{Dataset} & \multirow{2}{*}{Original Publication} & \multirow{2}{*}{Classification Task} & \multirow{2}{*}{No of Classes} & \multicolumn{2}{c}{Attributes} \\
    & & & & Attribute Type & \#N \\
    \hline
    \multirow{3}{*}{ECtHR} & \multirow{3}{*}{\cite{chalkidis-etal-2021-paragraph}} & \multirow{3}{*}{Legal Judgment Prediction: \emph{ECHR Violation Prediction}} & \multirow{3}{*}{10+1} & Defendant State & 2 \\
    & & & & Applicant Gender & 2 \\
    & & & & Applicant Age & 3 \\
    \hline
    \multirow{2}{*}{SCOTUS} & \multirow{2}{*}{\cite{spaeth2020}} & \multirow{2}{*}{Legal Topic Classification: \emph{Issue Area Classification}} & \multirow{2}{*}{14} & Respondent Type & 4 \\
    & & & & Decision Direction & 2 \\
    \hline
    \multirow{3}{*}{FSCS} & \multirow{3}{*}{\cite{niklaus-etal-2021-swiss}} & \multirow{3}{*}{Legal Judgment Prediction: \emph{Case Approval Prediction}} & \multirow{3}{*}{2} & Language & 3 \\
    & & & & Region of Origin & 6 \\
    & & & & Legal Area & 6 \\
    \hline
    \multirow{2}{*}{CAIL} & \multirow{2}{*}{\cite{wang-etal-2021-equality}} & \multirow{2}{*}{Legal Judgment Prediction: \emph{Crime Severity Prediction}} & \multirow{2}{*}{6} & Defendant Gender & 2 \\
    & & & & Region of Origin & 7 \\
    \end{tabular}
    }
    \vspace{-3mm}
    \caption{Main characteristics of FairLex datasets (ECtHR, SCOTUS, FSCS, CAIL). We report the examined tasks, the number of classes, the examined attributes and the number (\#N) of groups per attribute.}
    \label{tab:datasets}
    \vspace{-4mm}
\end{table*}

\paragraph{Benchmarking}

NLP has been stormed by the rapid development of benchmark datasets that aim to evaluate the performance of pre-trained language models with respect to different objectives: general Natural Language Understanding (NLU) \cite{wang-2019-glue,wang-2019-superglue}, Cross-Lingual Transfer (CLT)~\cite{hu2020xtreme}, and even domain-specific ones on biomedical \cite{peng2019transfer}, or legal \cite{chalkidis-2022-lexglue} NLP tasks. Despite their value, recent work has raised criticism on several limitations of the so called NLU benchmarks \cite{paullada-etal-2020-data, bowman-dahl-2021-will, raji2021ai}. The main points are: poor (\emph{laissez-faire}) dataset development (e.g., lack of diversity, spurious correlations), legal issues (e.g., data licensing and leakage of personal information), construct validity (e.g., poor experimental setup, unclear research questions), question of ``general'' capabilities, and promotion of superficial competitiveness (hype, or even falsify, state-of-the-art results).

We believe that the release of FairLex, a domain-specific (legal-oriented) benchmark suite  for evaluating fairness, overcomes (or at least mitigates) some of the aforementioned limitations. We introduce the core motivation in Section~\ref{sec:introduction}, while specific (case-by-case) details are described in Section~\ref{sec:datasets}. Our benchmark is open-ended and inevitably has several limitations; we report known limitations and ethical considerations in Sections~\ref{sec:limitations} and \ref{sec:ethics}. Nonetheless we believe that it will help critical research in the area of fairness.

\section{Benchmark Datasets}
\label{sec:datasets}
\paragraph{ECtHR} The European Court of Human Rights (ECtHR) hears allegations that a state has breached human rights provisions of the European Convention of Human Rights (ECHR). We use the dataset of \citet{chalkidis-etal-2021-paragraph}, which contains 11K cases from ECtHR's public database.
Each case is mapped to \emph{articles} of the ECHR that were violated (if any). This is a multi-label text classification task. Given the facts of a case, the goal is to predict the ECHR articles that were violated, if any, as decided (ruled) by the court. The cases are chronologically split into training (9k, 2001--16), development (1k, 2016--17), and test (1k, 2017--19) sets. 

To facilitate the study of fairness of text classifiers, we record for each case the following attributes: (a) The \emph{defendant states}, which are the European states that allegedly violated the ECHR. The defendant states for each case is a subset of the 47 Member States of the Council of Europe;\footnote{\url{https://www.coe.int/}} To have statistical support, we group defendant states in two: 
Central-Eastern European states, on one hand, and all other states, as classified by the EuroVoc thesaurus. (b) The \emph{applicant's age} at the time of the decision. We extract the birth year of the applicant from the case facts, if possible, and classify its case in an age group ($\leq$35, $\leq$64, or older) ; and (c) the \emph{applicant's gender}, extracted from the facts, if possible, based on pronouns or other gendered words, classified in two categories (male, female).\footnote{In Appendix~\ref{sec:attributes}, we describe attribute extraction and grouping in finer details for all datasets.}

\paragraph{SCOTUS}

The US Supreme Court (SCOTUS) is the highest federal court in the United States of America and generally hears only the most controversial or otherwise complex cases which have not been sufficiently well solved by lower courts.
We combine information from SCOTUS opinions with the Supreme Court DataBase (SCDB)\footnote{\url{http://scdb.wustl.edu}} \cite{spaeth2020}. SCDB provides metadata (e.g., date of publication, decisions, issues, decision directions and many more) for all cases. We consider the available 14 thematic issue areas (e.g, Criminal Procedure, Civil Rights, Economic Activity, etc.) as labels. This is a single-label multi-class document classification task. Given the court opinion, the goal is to predict the issue area whose focus is on the subject matter of the controversy (dispute). SCOTUS contains a total of 9,262 cases that we split chronologically into 80\% for training (7.4k, 1946--1982), 10\% for development (914, 1982--1991) and 10\% for testing (931, 1991--2016).

From SCDB, we also use the following attributes to study fairness: (a) the \emph{type of respondent}, which is a manual categorization of respondents (defendants) in five categories (person, public entity, organization, facility and other); and (c) the \emph{direction of the decision}, i.e., whether the decision is considered liberal, or conservative, provided by SCDB.

\paragraph{FSCS}
The Federal Supreme Court of Switzerland (FSCS) is the last level of appeal in Switzerland and similarly to SCOTUS, the court generally hears only the most controversial or otherwise complex cases which have not been sufficiently well solved by lower courts. The court often focus only on small parts of previous decision, where they discuss possible wrong reasoning by the lower court. The Swiss-Judgment-Predict dataset \cite{niklaus-etal-2021-swiss} contains more than 85K decisions from the FSCS written in one of three languages (50K German, 31K French, 4K Italian) from the years 2000 to 2020.
The dataset provides labels for a simplified binary (\emph{approval}, \emph{dismissal}) classification task. Given the facts of the case, the goal is to predict if the plaintiff's request is valid or partially valid. The cases are also chronologically split into training (59.7k, 2000-2014), development (8.2k, 2015-2016), and test (17.4k, 2017-2020) sets.

The original dataset provides three additional attributes: (a) the \emph{language} of the FSCS written decision, in either German, French, or Italian; (b) the \emph{legal area} of the case (e.g., public, penal law) derived from the chambers where the decisions were heard; and (c) the \emph{region} that denotes in which federal region was the case originated.

\paragraph{CAIL}
The Supreme People's Court of China is the last level of appeal in China and considers cases that originated from the high people's courts concerning matters of national importance. The Chinese AI and Law challenge (CAIL) dataset \cite{xiao-et-al-2018-cail} is a Chinese legal NLP dataset for judgment prediction and contains over 1m criminal cases. The dataset provides labels for \emph{relevant article of criminal code} prediction, \emph{charge} (type of crime) prediction, imprisonment \emph{term} (period) prediction, and monetary \emph{penalty} prediction.\footnote{The publication of the original dataset has been the topic of an active debate in the NLP community \cite{leins-etal-2020-give,tsarapatsanis-aletras-2021-ethical,bender2021}.}

Recently, \citet{wang-etal-2021-equality} re-annotated a subset of approx.\ 100k cases with demographic attributes. Specifically the new dataset has been annotated with: (a) the \emph{applicant's gender}, classified in two categories (male, female); and (b) the \emph{region} of the court that denotes in which out of the 7 provincial-level administrative regions was the case judged. We re-split the dataset chronologically into training (80k, 2013-2017), development (12k, 2017-2018), and test (12k, 2018) sets. In our study, we re-frame the imprisonment \emph{term} prediction and examine a soft version, dubbed \emph{crime severity} prediction task, a multi-class classification task, where given the facts of a case, the goal is to predict how severe was the committed crime with respect to the imprisonment term. We approximate crime severity by the length of imprisonment term, split in 6 clusters (0, $\leq$12, $\leq$36, $\leq$60, $\leq$120, >120 months).

\section{Fine-tuning Algorithms}
\label{sec:algorithms}

Across experiments, our main goal is to find a hypothesis for which the risk $R(h)$ is minimal:\vspace{-2mm}
\begin{flalign}
h^{*} &= \arg\min_{{h\in{\mathcal{H}}}}R(h) \\
R(h) &= \EX(L(h(x),y))
\end{flalign}

\noindent where $y$ are the targets (\emph{ground truth}) and $h(x) = \hat{y}$ is the system hypothesis (model's predictions).\vspace{2mm}

Similar to previous studies, $R(h)$ is an expectation of the selected loss function ($\mathcal{L}$). In this work, we study multi-label text classification (Section~\ref{sec:datasets}), thus we aim to minimize the binary cross-entropy loss across $L$ classes:\vspace{-2mm}
\begin{equation}
   \mathcal{L} =\ -y\log {\hat {y}}-(1-y)\log(1-{\hat {y}})
\end{equation}

\noindent\textbf{ERM}~\cite{vapnik-1992}, which stands for Empirical Risk Minimization, is the most standard and widely used optimization technique to train neural methods. The loss is calculated as follows:\vspace{-2mm}
\begin{equation}
    \mathcal{L}_{ERM} = \sum_{i=1}^{N} \frac{\mathcal{L}_i}{N}
\end{equation}

\noindent where $N$ is the number of instances (training examples) in a batch, and $\mathcal{L}_i$ is the loss per instance.

Besides ERM, we also consider a representative selection of group-robust fine-tuning algorithms which aims at mitigating performance disparities with respect to a given attribute ($A$), e.g., the gender of the applicant or the region of the court. Each attribute is split into $G$ groups, i.e., male/female for gender. All algorithms rely on a balanced group sampler, i.e., an equal number of instances (samples) per group ($N_G$) are included in each batch. Most of the algorithms are built upon group-wise losses ($\mathcal{L}_g$), computed as follows:\vspace{-2mm}

\begin{equation}
    \mathcal{L}(g_i) = \frac{1}{N_{g_i}}\sum_{j=1}^{N_{g_i}} \mathcal{L}(x_j)
\end{equation}

\noindent\textbf{Group DRO}~\cite{sagawa-etal-2020-dro}, stands for Group Distributionally Robust Optimization (DRO). Group DRO is an extension of the Group Uniform algorithm, where the group-wise losses are weighted inversely proportional to the group training performance. The total loss is:
\begin{equation}
    \mathcal{L}_{DRO} = \sum_{i=1}^{G}w_{g_i} * \mathcal{L}(g_i)\textrm{, where}
    \label{eq:dro_loss}
\end{equation}
\vspace{-6mm}
\begin{equation}
      w_{g_i} = \frac{1}{W}(\hat{w}_{g_i} * e^{L(g_i)}) \quad \textrm{and} \quad W = \sum_{i=1}^{G} w_{g_i}
    \label{eq:dro_loss_w}
\end{equation} 

\noindent where $G$ is the number of groups (labels), $\mathcal{L}_g$ are the averaged group-wise (label-wise) losses, $w_g$ are the group (label) weights, $\hat{w_g}$ are the group (label) weights as computed in the previous update step. Initially the weight mass in equally distributed across groups.\vspace{2mm}

\noindent\textbf{V-REx}~\cite{krueger-etal-2020-rex}, which stands for Risk Extrapolation, is yet another proposed group-robust optimization algorithm. \citet{krueger-etal-2020-rex} hypothesize that variation across training groups is representative of the variation later encountered at test time, so they also consider the variance across the group-wise losses. In V-REx the total loss is calculated as follows:\vspace{-2mm}
\begin{equation}
    \mathcal{L}_{REX} = \mathcal{L}_{ERM} + \lambda * \mathrm{Var}([\mathcal{L}_{g_1},\dots,\mathcal{L}_{g_G}])
\end{equation}

\noindent where $\mathrm{Var}$ is the variance among the group-wise losses and $\lambda$, a weighting hyper-parameter scalar.\vspace{2mm}

\noindent\textbf{IRM}~\cite{arjovsky-etal-2020-irm}, which stands for Invariant Risk Minimization, mainly aims to penalize variance across multiple training dummy estimators across groups, i.e., performance cannot vary in samples that correspond to the same group. The total loss is computed as follows:\vspace{-2mm}
\begin{equation}
    \mathcal{L}_{IRM} = \frac{1}{G} \left(\sum_{i=1}^{G} \mathcal{L}(g_i) + \lambda * P(g_i)\right)
     \label{eq:irm_loss}
\end{equation}

Please refer to~\citet{arjovsky-etal-2020-irm} for the definition of the group penalty terms ($P_{g}$).\vspace{2mm}

\noindent\textbf{Adversarial Removal}~\cite{elazar-goldberg-2018-adversarial} algorithm mitigates group disparities by means of an additional adversarial classifier \cite{goodfellow2014generative}. The adversarial classifier share the encoder with the main network and is trained to predict the protected attribute ($A$) of an instance. The total loss factors in the adversarial one, thus penalizing the model when it is able to discriminate groups.
Formally, the total loss is calculated as:
\begin{equation}
    \mathcal{L}_{AR} = \mathcal{L}_{ERM} - \lambda * \mathcal{L}_{ADV}
\end{equation}
\vspace{-4mm} 
\begin{equation}
    \mathcal{L}_{ADV} = \mathcal{L}(\hat{g}_i, g_i)
\end{equation}

\noindent where $\hat{g}_i$ is the adversarial classifier's prediction for the examined attribute A (in which group ($g_i$) of $A$, does the example belong to) given the input ($x$).

\section{Experimental Setup}

\paragraph{Models} 
\label{sec:models}
Since we are interested in classifying long documents (up to 6000 tokens per document, see Figure~\ref{fig:datasets} 
in Appendix~\ref{sec:appendix_b}), we use a hierarchical BERT-based model similar to that of \citet{chalkidis-etal-2021-paragraph}, so as to avoid using only the first 512 tokens of a text. 
The hierarchical model, first, encodes the text through a pre-trained Transformer-based model, thus representing each paragraph independently with the [CLS] token. Then, the paragraph representations are fed into a two-layers transformer encoder with the exact same specifications of the first one (e.g., hidden units, number of attention heads), so as to contextualize them, i.e., it makes paragraphs representations aware of the surrounding paragraphs. Finally, the model max-pools the context-aware paragraph representations computing the document-level representation and feed it to a classification layer.

For the purpose of this work, we release four domain-specific BERT models with continued pre-training on the corpora of the examined datasets.\footnote{\url{https://huggingface.co/coastalcph}} We train mini-sized BERT models with 6 Transformer blocks, 384 hidden units, and 12 attention heads. We warm-start all models from the public MiniLMv2 models checkpoints \cite{wang-etal-2021-minilmv2} using the distilled version of RoBERTa~\cite{liu-etal-2019-roberta} for the English datasets (ECtHR, SCOTUS) and the one distilled from XLM-R~\cite{conneau-etal-2020-unsupervised} for the rest (trilingual FSCS, and Chinese CAIL). Given the limited size of these models, we can effectively use up to 4096 tokens in ECtHR and SCOTUS and up to 2048 tokens in FSCS and CAIL for up to 16 samples per batch in a 24GB GPU card.\footnote{This is particularly important for group-robust algorithms that consider group-wise losses.} 
For completeness, we also consider linear Bag-of Words (BoW) classifiers using TF-IDF scores of the most frequent $n$-grams (where $n=1,2,3$) in the training corpus of each dataset.

\paragraph{Data Repository and Code} We release a unified version of the benchmark on Hugging Face Datasets~\cite{lhoest2021datasets}.\footnote{\url{https://huggingface.co/datasets/coastalcph/fairlex}} In our experiments, we use and extend the WILDs \cite{wilds2021} library. For reproducibility and further exploration with new group-robust methods, we release our code on Github.\footnote{\url{https://github.com/coastalcph/fairlex}}

\paragraph{Evaluation Details} 
Across experiments we compute the macro-F1 score per group ($\mathrm{mF1}_i$), excluding the group of \emph{unidentified} instances, if any.\footnote{The group of \emph{unidentified} instances includes the instances, where the value of the examined attribute is unidentifiable (unknown). See details in Appendices~\ref{sec:attributes}, and \ref{sec:group_dist}.} We report macro-F1 to avoid bias toward majority classes because of class imbalance and skewed label distributions across train, development, and test subsets \cite{sogaard-etal-2021-need}.

\section{Results}

\begin{table}[]
    \centering
    \vspace{-20mm}
    \resizebox{\columnwidth}{!}{
    \begin{tabular}{l|r|r|r|r}
        \hline \hline
         \multicolumn{5}{c}{\bf ECtHR (ECHR Violation Prediction)} \\
         \hline \hline
         Group &  \multicolumn{1}{c|}{mF1} & \multicolumn{1}{c|}{\#train-cases (\%) ($\uparrow$)} & \multicolumn{1}{c|}{$LD_{KL}$ ($\downarrow$)} & \multicolumn{1}{c}{$\mathrm{WCI}$ ($\downarrow$)}  \\
         \hline
         \multicolumn{5}{c}{\textsc{Defendant State}} \\
         \hline
         \cellcolor{best} \bf E.C. European & \cellcolor{best} \bf 70.2 & \cellcolor{best} \bf 7,224 (80\%) & \cellcolor{best} \bf 0.17 & \cellcolor{best} \bf 0.07 \\
         \cellcolor{worst} \bf \emph{The Rest} & \cellcolor{worst} 48.7 & \cellcolor{worst} 1,776 (20\%) & \cellcolor{worst} 0.28 & \cellcolor{worst} 0.57 \\
         \hline
         \multicolumn{5}{c}{\textsc{Applicant Gender}} \\
         \hline
         \cellcolor{worst} \bf Male & \cellcolor{worst} 54.4 & \cellcolor{worst} \bf 4,187 (77\%) & \cellcolor{worst} \bf 0.17 & \cellcolor{worst} \bf 0.18 \\
         \cellcolor{best} \bf \emph{Female} & \cellcolor{best} \bf 60.6 & \cellcolor{best} 1,507 (23\%) & \cellcolor{best} 0.26 & \cellcolor{best} 0.19 \\
         \hline
         \multicolumn{5}{c}{\textsc{Applicant Age}} \\
         \hline
         \cellcolor{best} \bf $\leq\textrm{65}$ years & \cellcolor{best} \bf 59.7 & \cellcolor{best} \bf 4279 (68\%) & \cellcolor{best} \bf 0.18 &\cellcolor{best}  0.15 \\
          > 65 years &  56.5 &  1130 (18\%) &  0.32 & 0.26 \\
         \cellcolor{worst} \bf \emph{$\leq\textrm{35}$ years} & \cellcolor{worst} 46.2 & \cellcolor{worst} 868 (14\%) & \cellcolor{worst} 0.19 & \cellcolor{worst} \bf 0.12 \\
         \hline
         \multicolumn{5}{c}{}\\
         \hline \hline
         \multicolumn{5}{c}{\bf SCOTUS (Issue Area Classification)} \\
         \hline \hline
         Group &  \multicolumn{1}{c|}{mF1} & \multicolumn{1}{c|}{\#train-cases (\%) ($\uparrow$)}& \multicolumn{1}{c|}{$LD_{KL}$ ($\downarrow$)} & \multicolumn{1}{c}{$\mathrm{WCI}$ ($\downarrow$)}   \\
         \hline
         \multicolumn{5}{c}{\textsc{Respondent Type}} \\
         \hline
         Public Entity & 77.4 & \bf 2796 (51\%) & 0.07 & 0.04 \\
         \cellcolor{worst} \bf \emph{Person} & \cellcolor{worst} 74.9 & \cellcolor{worst} 1847 (34\%) & \cellcolor{worst} \bf 0.05 & \cellcolor{worst} \bf 0.03 \\
         \cellcolor{best} \bf Organization & \cellcolor{best} \bf 81.1 & \cellcolor{best} 741 (13\%) & \cellcolor{best} 0.11 & \cellcolor{best} \bf 0.03 \\
         Facility & 80.7 & 140 (3\%) & 0.26 & 0.06 \\
         \hline
         \multicolumn{5}{c}{\textsc{Direction}} \\
         \hline
         \cellcolor{worst} \bf \emph{Liberal} & \cellcolor{worst} 76.2 & \cellcolor{worst} \bf 3335 (52\%) & \cellcolor{worst} \bf 0.04 &  \cellcolor{worst} \bf 0.08 \\
         \cellcolor{best} \bf Conservative & \cellcolor{best} \bf 80.8 & \cellcolor{best} 3146 (48\%) & \cellcolor{best} 0.05 & \cellcolor{best} 0.17 \\
         \hline
         \multicolumn{5}{c}{}\\
         \hline \hline
         \multicolumn{5}{c}{\bf FSCS (Case Approval Prediction)} \\
         \hline \hline
         Group & \multicolumn{1}{c|}{mF1} & \multicolumn{1}{c|}{\#train-cases (\%) ($\uparrow$)}& \multicolumn{1}{c|}{$LD_{KL}$ ($\downarrow$)} &  \multicolumn{1}{c}{$\mathrm{WCI}$ ($\downarrow$)}  \\
         \hline
         \multicolumn{5}{c}{\textsc{Language}} \\
         \hline
         German &  68.2 &  \bf 35458 (60\%) & \bf  0.03 &  0.20 \\
         \cellcolor{best} \bf French & \cellcolor{best} \bf 70.6 & \cellcolor{best} 21179 (35\%) & \cellcolor{best} \bf  0.03 & \cellcolor{best} \bf 0.19 \\
         \cellcolor{worst} \bf \emph{Italian} & \cellcolor{worst} 65.2 & \cellcolor{worst} 3072 (5\%) & \cellcolor{worst} 0.04 & \cellcolor{worst} \bf 0.19 \\
         \hline
         \multicolumn{5}{c}{\textsc{Legal Area}} \\
         \hline
         \cellcolor{worst} \bf \emph{Public law} & \cellcolor{worst} 56.9 & \cellcolor{worst} \bf  15173 (31\%) & \cellcolor{worst} \bf  \textasciitilde0.00 & \cellcolor{worst} 0.20 \\
         \cellcolor{best} \bf Penal law & \cellcolor{best} \bf 83.4 & \cellcolor{best} 11795 (25\%) & \cellcolor{best} \bf \textasciitilde0.00 & \cellcolor{best} 0.20 \\
         Civil law & 66.4 &  11477 (24\%) & 0.02 &  \bf  0.16 \\
         Social Law & 70.8 & 9727 (20\%) &  0.06 & 0.20 \\
         \hline
         \multicolumn{5}{c}{\textsc{Region}} \\
         \hline
         R. Lémanique & 71.3 & \bf 13436 (27\%) & 0.04 & 0.20 \\
         Zürich & 68.5 &  8788 (18\%) & 0.04 & 0.18 \\
         E. Mittelland & 69.8 & 8257 (17\%) & 0.08 & \bf 0.16 \\
         \cellcolor{best} \bf E. Switzerland & \cellcolor{best} \bf 73.6 & \cellcolor{best} 5707 (12\%) & \cellcolor{best} 0.02 & \cellcolor{best} 0.24 \\
         N.W. Switzerland & 72.8 & 5655 (11\%) & 0.03 & 0.19 \\
         C. Switzerland & 69.5 & 4779 (10\%) & 0.03 & 0.19 \\
         Ticino & 68.3 & 2255 (6\%) & \bf \textasciitilde0.00 & 0.17 \\
         \cellcolor{worst} \bf \emph{Federation} & \cellcolor{worst} 63.9 & \cellcolor{worst} 1308 (3\%) & \cellcolor{worst} \bf \textasciitilde0.00 & \cellcolor{worst} 0.27 \\
         \multicolumn{5}{c}{}\\
         \hline \hline
         \multicolumn{5}{c}{\bf CAIL (Crime Severity Prediction)} \\
         \hline \hline
         Group & \multicolumn{1}{c|}{mF1} & \multicolumn{1}{c|}{\#train-cases (\%) ($\uparrow$)}& \multicolumn{1}{c|}{$LD_{KL}$ ($\downarrow$)} &  \multicolumn{1}{c}{$\mathrm{WCI}$ ($\downarrow$)}  \\
         \hline
         \multicolumn{5}{c}{\textsc{Defendant Gender}} \\
         \hline
         \cellcolor{best} \bf \emph{Male}  & \cellcolor{best} 60.3 & \cellcolor{best} \bf 73952 (92\%) & \cellcolor{best} \bf 0.03 & \cellcolor{best} 0.01 \\
         \cellcolor{worst} \bf Female & \cellcolor{worst} 60.1 & \cellcolor{worst} 6048 (8\%) & \cellcolor{worst} 0.08 & \cellcolor{worst} \bf 0.03 \\
         \hline
         \multicolumn{5}{c}{\textsc{Region}} \\
         \hline
         \cellcolor{best} \bf Beijing & \cellcolor{best} \bf 66.8 & \cellcolor{best} \bf 16588 (21\%) & \cellcolor{best} \bf 0.05 & \cellcolor{best} \bf 0.02 \\
         Liaoning & 56.7 & 13934 (17\%) & \bf 0.05 & \bf 0.02 \\
         Hunan & 59.5 & 12760 (16\%) & \bf 0.05 & \bf 0.02 \\
         Guangdong & 58.0 & 12278 (15\%) & \bf 0.05 & 0.01 \\
         \cellcolor{worst} \bf \emph{Sichuan} & \cellcolor{worst} 56.4 & \cellcolor{worst} 11606 (14\%) & \cellcolor{worst} 0.06 & \cellcolor{worst} \bf 0.02 \\
         Guangxi & 58.9 &  8674 (11\%) & 0.07 & \bf 0.02 \\
         Zhejiang & 58.8 & 4160 (5\%) & 0.07 & \bf 0.02 \\
         \hline 

    \end{tabular}
    }
    \caption{Statistics for the three general (attribute agnostic) cross-examined factors (\emph{representation inequality}, \emph{temporal concept drift}, and \emph{worst-class influence}), as introduced in Section~\ref{sec:analysis}. We highlight the \colorbox{worst}{\emph{\textbf{worst}}} and \colorbox{best}{\textbf{best}} performing group per attribute. In \textbf{boldface}, we highlight the best (less harmful) value per factor across groups. Performance (mF1) reported for ERM.}
    \label{tab:factors}
    \vspace{-20mm}
\end{table}

\paragraph{Main Results}

In Table~\ref{tab:factors}, we report the group performance (mF1), where models trained with the ERM algorithm, across all datasets and attributes. We observe that the intensity of group disparities vary a lot between different attributes, but in many cases the group disparities are very vibrant.

For example, in ECtHR, we observe substantial group disparity between the two \emph{defendant state} groups (21.5\% absolute difference), similarly for \emph{applicant's gender} groups (16.2\% absolute difference). In FSCS, we observe \emph{language} disparity, where performance is on average 3-5\% lower for cases written in Italian compared to those written in French and German. Performance disparity is even higher with respect to \emph{legal areas}, where the model has the best performance for criminal (penal law) cases  (83.4\%) compared to others (approx.\ 10-20\% lower). We also observe substantial group disparities with respect to the \emph{court region}, e.g., cases ruled in E. Switzerland courts (66.8\%) compared to Federation courts (56.4\%). The same applies for CAIL, e.g., cases ruled in Beijing courts (66.8\%) compared to Sichuan courts (56.4\%).

\paragraph{Group Disparity Analysis}
\label{sec:analysis} Moving forward we try to identify general (attribute agnostic) factors based on data distributions that could potentially lead to performance disparity across groups. We identify three general (attribute agnostic) factors:\vspace{-1mm}

\begin{itemize}[leftmargin=8pt]
    \item \emph{Representation Inequality}: Not all groups are equally represented in the training set. To examine this aspect, we report the number of training cases per group.
    \item \emph{Temporal Concept Drift}: The label distribution for a given group changes over time, i.e., in-between training and test subsets. To examine this aspect, we report per group, the KL divergence in-between the training and test label distribution.
    \item \emph{Worst Class Influence}: The performance is not equal across labels (classes), which may dis-proportionally affect the macro-averaged performance across groups. To examine this aspect, we report the \emph{Worst Class Influence (WCI)} score per group, which is computed as follows:\vspace{-1mm}
    \begin{equation}
        \mathrm{WCI}(i) = \frac{\textrm{\#test-cases (worst-class)}}{\textrm{\#test-cases}}
    \end{equation}
\end{itemize}  

In Table~\ref{tab:factors}, we present the results across all attributes. We observe that only in 4 out of 10 cases (attributes), the less represented groups are those with the worst performance compared to the rest. It is generally not the case that high KL divergence (drift) correlates with low performance. In other words, group disparities does not seem to be driven by temporal concept drift. Finally, the influence of the worst class is relatively uniform across groups in most cases, but in the cases where groups differ in this regard, worst class influence correlates with error in 2 out of 3 cases.\footnote{For ECtHR performance across defendant states and SCOTUS across directions, but not for ECtHR performance across applicant age.}

In ECtHR, considering performance across defendant state, we see that all the three factors correlate internally, i.e., the worst performing group is less represented, has higher temporal drift and has more cases in the worst performing class. This is not the case considering performance across other attributes. It is also not the case for SCOTUS. 
In FSCS, considering the attributes of language and region, representation inequality seems to be an important factor that leads to group disparity. This is not the case for legal area, where the best represented group is the worst performing group. In other words, there are other reasons that lead to performance disparity in this case; 
according to \citet{niklaus-etal-2021-swiss}, a potential factor is that the jurisprudence in penal law is more united and aligned in Switzerland and outlier judgments are rarer making the task more predictable.

\begin{table}[t]
    \centering
    \resizebox{0.9\columnwidth}{!}{
    \begin{tabular}{l|ccc}
        \hline
        \multicolumn{4}{c}{\bf ECtHR ($A_1$: Defendant State)} \\
        \hline
         Group ($A_2$)  & E.C.E.  & Rest & Avg. \\
        \midrule 
        \cellcolor{worst} Male  & \cellcolor{worst} 55.8   &\cellcolor{worst}  35.1  & \cellcolor{worst} 54.4 \\
        \cellcolor{best} Female  & \cellcolor{best} \bf 61.3   & \cellcolor{best} \bf 47.1  & \cellcolor{best} \bf 60.6 \\
        \hline
        \cellcolor{worst} $\leq$35  & \cellcolor{worst} 48.1   & \cellcolor{worst} \bf 44.2  & \cellcolor{worst} 46.2 \\
        \cellcolor{best} $\leq$65  & \cellcolor{best} \bf 61.0 & \cellcolor{best} 34.7  & \cellcolor{best} \bf 59.7 \\
        \hline
        \multicolumn{4}{c}{\bf FSCS ($A_1$: Legal Area)} \\
        \hline
        Group ($A_2$)   & Public Law  & Penal Law  & Avg. \\
        \midrule 
        \cellcolor{best} French    & \cellcolor{best} \bf 57.4    & \cellcolor{best} \bf 82.4   & \cellcolor{best} \bf 70.6 \\
        \cellcolor{worst} Italian   & \cellcolor{worst} 56.2    & \cellcolor{worst} 69.4   & \cellcolor{worst} 65.2 \\
        \hline
        \cellcolor{best} E. Switzerland  & \cellcolor{best} \bf 55.9    & \cellcolor{best} \bf 87.0   & \cellcolor{best} \bf 73.6 \\
        \cellcolor{worst} Federation   & \cellcolor{worst} 54.5    & \cellcolor{worst} 72.8   &\cellcolor{worst} 63.9 \\
        \hline
    \end{tabular}
    }
    \vspace{-2mm}
    \caption{Results in cross-attribute influence. mF1 scores for pairings of groups for attributes ($A_1$, $A_2$).}
    \vspace{-6mm}
    \label{tab:cross-attribute}
\end{table}

\paragraph{Cross-Attribute Influence Analysis} We have evaluated fairness across attributes that are not necessarily independent of each other. We therefore evaluate the extent to which performance disparities along different attributes correlate, i.e., how attributes interact, and whether performance differences for attribute $A_1$ can potentially explain performance differences for another attribute $A_2$. 
We examine this for the two attributes with the highest group disparity: the \emph{defendant state} in ECtHR, and the \emph{legal area} in FSCS. For the bins induced by these two attributes ($A_1$), we compute mF1 scores across other attributes ($A_2$).

\begin{table*}[t]
    \centering
    \resizebox{\textwidth}{!}{
    \begin{tabular}{l|ccc|ccc|ccc||ccc|ccc}
        \hline
        \hline
          & \multicolumn{9}{c||}{\bf ECtHR (ECHR Violation Prediction)} & \multicolumn{6}{c}{\bf SCOTUS (Issue Area Classification)} \\
        \hline
        \hline
         \bf \multirow{2}{*}{Algorithm} & \multicolumn{3}{c|}{\emph{Defendant State}} & \multicolumn{3}{c|}{\emph{Applicant Gender}} & \multicolumn{3}{c||}{\emph{Applicant Age}} & \multicolumn{3}{c|}{\emph{Respondent Type}} & \multicolumn{3}{c}{\emph{Direction}} \\
         & $\uparrow\overline{\mathrm{mF1}}$ & $\downarrow\mathrm{GD}$ & $\uparrow{\mathrm{mF1}_W}$ & $\uparrow\overline{\mathrm{mF1}}$ & $\downarrow\mathrm{GD}$ & $\uparrow{\mathrm{mF1}_W}$ & 
         $\uparrow\overline{\mathrm{mF1}}$ & $\downarrow\mathrm{GD}$ & $\uparrow{\mathrm{mF1}_W}$ & $\uparrow\overline{\mathrm{mF1}}$ & $\downarrow\mathrm{GD}$ & $\uparrow{\mathrm{mF1}_W}$ & $\uparrow\overline{\mathrm{mF1}}$ & $\downarrow\mathrm{GD}$ & $\uparrow{\mathrm{mF1}_W}$  \\
         \hline
         \multicolumn{16}{c}{\textsc{Bag-of-Words Linear Classifier}} \\
         \hline
         ERM & 46.8 & 3.0 & 43.8 & 44.1 & 4.9 & 40.6 & 46.9 & 6.3 & 40.9 & 73.8 & 6.6 & 61.8 & 77.5 & 2.6 & 74.9 \\
         \hline
         \multicolumn{16}{c}{\textsc{Transformer-based Classifier}} \\
         \hline
         ERM & 53.2 & 8.3 & 44.9 & 57.5 & 3.1 & 54.4 & 54.1 & 5.9 & 46.2 & 75.1 & 4.0 & 70.8 & 78.1 & 1.6 & 76.6  \\
         ERM+GS & 54.4 & 5.5 & 48.9 & \bf 57.8 & 3.3 & 54.5  & \bf 56.0 & 5.6 & 48.7 & \bf 75.2 & 3.9 & 70.9 & 77.1 & 1.3 & 76.0 \\
         \hline
        ADV-R & 53.8 & 5.8 & 47.9 & 54.6 & 3.2 & 51.5 & 48.9 & 6.1 & 40.6 & 56.9 & 4.7 & 53.1 & 41.0 & \bf 0.8 & 40.3  \\
         G-DRO & \bf 55.0 & \bf 5.2 & \bf 49.8 & 56.3 & \bf 1.9 & \bf 55.0 & 52.6 & 6.2 & 44.3 & 74.5 & 3.3 & \bf 71.6 & 77.1 & 1.7 & 75.4 \\
         IRM & 53.8 & 5.7 & 48.1 & 53.8 & 2.3 & 52.5 & 54.8 & \bf 4.4 & 49.5 & 73.4 & 4.8 & 68.2 & 78.1 & 2.7 & 75.4  \\
         V-REx & 54.6 & 6.3 & 48.3 & 54.6 & 2.0 & 53.2 & 55.0 & 4.5 & \bf 49.8 & 73.8 & \bf 3.8 & 68.2 & \bf 78.2 & 1.1 & \bf 77.1  \\
         \hline
        \hline
          & \multicolumn{9}{c||}{\bf FSCS (Case Approval Prediction)} & \multicolumn{6}{c}{\bf CAIL (Crime Severity Prediction)} \\
         \hline
        \hline
         \bf \multirow{2}{*}{Algorithm} & \multicolumn{3}{c|}{\emph{Language}} & \multicolumn{3}{c|}{\emph{Legal Area}} & \multicolumn{3}{c||}{\emph{Region}} & \multicolumn{3}{c|}{\emph{Defendant Gender}} & \multicolumn{3}{c}{\emph{Region}}\\ 
          & $\uparrow\overline{\mathrm{mF1}}$ & $\downarrow\mathrm{GD}$ & $\uparrow{\mathrm{mF1}_W}$ & $\uparrow\overline{\mathrm{mF1}}$ & $\downarrow\mathrm{GD}$ & $\uparrow{\mathrm{mF1}_W}$ & 
         $\uparrow\overline{\mathrm{mF1}}$ & $\downarrow\mathrm{GD}$ & $\uparrow{\mathrm{mF1}_W}$ & $\uparrow\overline{\mathrm{mF1}}$ & $\downarrow\mathrm{GD}$ & $\uparrow{\mathrm{mF1}_W}$ & $\uparrow\overline{\mathrm{mF1}}$ & $\downarrow\mathrm{GD}$ & $\uparrow{\mathrm{mF1}_W}$ \\
         \hline
         \multicolumn{16}{c}{\textsc{Bag-of-Words Linear Classifier}} \\
         \hline
         ERM & 55.5 & 6.2 & 46.8 & 54.4 & 9.7 & 40.9 & 56.8 & 5.0 & 46.6 & 33.5 & 0.7 & 32.8 & 31.7 & 5.0 & 25.5 \\
         \hline
         \multicolumn{16}{c}{\textsc{Transformer-based Classifier}} \\
         \hline
         ERM & 67.8 & 2.1 & 65.0 & \bf 69.4 & 9.6 & \bf 56.9 & \bf 69.7 & \bf 2.9 & \bf 63.9 & \bf 60.2 & \bf 0.6 & \bf 60.1 & \bf 59.3 & 3.5 & \bf 56.4 \\
         ERM+GS & 66.4 & 3.5 & 61.7 & 67.1 & 9.3 & 55.5 & 67.9 & 3.0 & 62.3 & 59.4 & 0.7 & 59.1 & 58.2 & 3.1 & 55.9 \\
         \hline
         ADV-R & 62.6 & 5.1 & 59.0 & 65.6 & 12.4 & 50.0 & 67.4 & 3.2 & 61.5 & 53.3 & 1.3 & 52.1 & 53.5 & \bf 2.5 & 50.8 \\
         G-DRO & \bf 70.5 & \bf 0.6 & \bf 69.9 & 57.5 & \bf 5.6 & 52.6 & 67.7 & 4.2 & 60.2 & 59.2 & 1.3 & 57.9 & 58.9 & 3.7 & 55.7 \\
         IRM & 68.3 & 1.9 & 66.7 & 67.8 & 9.5 & 55.8 & 68.7 & 3.0 & 63.2 & 56.4 & 1.5 & 55.7 & 58.0 & 3.1 & 54.9 \\
         V-REx & 67.2 & 3.5 & 62.4 & 66.6 & 8.9 & 56.0 & 68.4 & 3.1 & 62.4 & 58.5 & 0.7 & 58.3 & 58.6 & 3.3 & 54.4 \\
         \hline
    \end{tabular}
    }
    \vspace{-2mm}
    \caption{Test results for all examined group-robust algorithms per dataset attribute. We report the average performance across groups ($\overline{\mathrm{mF1}}$), the \emph{group disparity} among groups ($\mathrm{GD}$), and the worst-group performance (${\mathrm{mF1}_W}$). $\uparrow$ denotes that higher scores are better, while $\downarrow$ denotes that lower scores are better.}
    \label{tab:baselines}
    \vspace{-5mm}
\end{table*}

In ECtHR, approx.\ 83\% and 81\% of \emph{male} and \emph{women} applicants are involved in cases against \emph{E.C. European} states (best-performing group). Similarly, in case of age groups, we observe that ratio of cases against E.C. European states  is: 87\% and 86\% for $\leq$65 and $\leq$35, the best- and worst-performing groups respectively. 
In FSCS, the ratio of cases relevant to \emph{penal law} is:  approx.\ 29\%, and 41\% written in written in \emph{French} (best-performing group) and \emph{Italian} (worst-performing group). Similarly, approx.\ 27\% originated in \emph{E. Switzerland} (best-performing group) and 42\% in \emph{Federation} (worst performing group) are relevant to public law. In both attributes, there is a 15\% increase of cases relevant to public law for the worst performing groups. In other words, the group disparity in one attribute $A_2$ (language, region) could be also explained by the influence of another attribute $A_1$ (legal area).

In Table~\ref{tab:cross-attribute}, we report the performance in the aforementioned cross-attribute ($A_1$, $A_2$) pairings. With the exception of the (age, defendant state) cross-examination in ECtHR, we observe that group disparities in attribute $A_2$ (Table~\ref{tab:factors}) are consistent across groups of the plausible influencer (i.e. attribute $A_1$). Hence, cross-attribute influence does not explain the observed group disparities.

We believe that such an in-depth analysis of the results is fundamental to understand the influence of different factors in the outcomes. This analysis wouldn't be possible, if we had ``counterfeited'' an ideal scenario, where all groups and labels where equally represented. While a controlled experimental environment is frequently used to examine specific factors, it could hide, or partially alleviate such phenomena, hence producing misleading results on fairness of the examined models.

\paragraph{Group Robust Algorithms Results} Finally, we evaluate the performance for several group robust algorithms ( Section~\ref{sec:algorithms}) that could potentially mitigate group disparities. To estimate their performance,  we report the average macro-F1 across groups ($\overline{\mathrm{mF1}}$) and the \emph{group disparity} ($\mathrm{GD}$) among groups measured as the group-wise std dev.:
\begin{equation}
    GD = \sqrt{\frac{1}{G} \sum_{i=1}^G (\mathrm{mF1}_i - \overline{\mathrm{mF1}})^2}
\end{equation}\vspace{-1mm}

We also report the \emph{worst-group performance} ($\mathrm{mF1}_W = \min([\mathrm{mF1}_1, \mathrm{mF1}_2, \dots \mathrm{mF1}_G)$).

In Table~\ref{tab:baselines}, we report the results of all our baselines on the four datasets introduced in this paper. 
We first observe that the results of linear classifiers trained with the ERM algorithm (top row per dataset) are consistently worse (lower average and worst-case performance, higher group disparity) compared to transformed-based models in the same setting. In other words linear classifier have lower overall performance, while being less \emph{fair} with respect to the applied definition of fairness (i.e. equal performance across groups).

As one can see, transformer-based models trained with the ERM algorithm, i.e., without taking into account information about groups and their distribution, perform either better on in the same ballpark than models trained with methods specialized to mitigate biases (Section~\ref{sec:algorithms}), with an average loss of \num{0.17}\% only in terms of $mF1$ and of \num{0.78}\% in terms of $mF1_W$.
While, these algorithms improve worst case performance in the literature, when applied in a controlled experimental environment, they fail in a more realistic setting, where both groups across attributes and labels are imbalanced, while also both group and label distribution change over time. Furthermore, we cannot identify one algorithm that performs better across datasets and group with respect to the others, indeed results are quite mixed without any recognizable pattern.

\section{Limitations}
\label{sec:limitations}

The current version of FairLex covers a very small fraction of legal applications, jurisdictions, and protected attributes. Our benchmark is open-ended and inevitably cannot cover ``\emph{everything in the whole wide (legal) world}'' \cite{raji2021ai}, but nonetheless we believe that the published resources will help critical research in the area of fairness. Some protected attributes within our datasets are extracted automatically, i.e., the gender and the age in the ECtHR dataset, if possible, by means of regular expressions, or manually clustered by the authors, such as the defendant state in the ECtHR dataset and the respondent attribute in the SCOTUS dataset. Various simplifications made, e.g, the binarization of gender, would be inappropriate in real-world applications. 

Another important limitation is that what is considered the {\em ground truth} in these datasets (with the exception of SCOTUS) is only ground truth relative to judges' interpretation of a specific (EC, US, Swiss, Chinese) jurisdiction and legal framework. The labeling is therefore somewhat subjective for non-trivial cases, and its validity is only relative to a given legal framework. We of course do not in any way endorse the legal standards or framework of the examined datasets.

\section{Conclusions}

We introduced FairLex, a multi-lingual benchmark suite for the development and testing of models and bias-mitigation algorithms within the legal domain, based on four datasets covering four jurisdictions, five languages and various sensitive attributes. Furthermore, we provided competitive baselines including transformer-based language models adapted to the examined datasets, and examination of performance of four group robust algorithms (Adversarial Removal, IRM, Group DRO, and V-REx). While, these algorithms improve worst case performance in the literature, when applied in a controlled experimental environment, they fail in a more realistic setting, where both groups across attributes, and labels are imbalanced, while also both group and label distributions change over time. Furthermore, we cannot identify a single algorithm that performs better across datasets and groups compared to the rest. 

In future work, we aim to further expand the benchmark with more datasets that could possibly cover more sensitive attributes. Further analysis on the reasons behind group disparities, e.g., representational bias, systemic bias, is also critical.

\section*{Ethics Statement}
\label{sec:ethics}

\noindent\textbf{Social Impact of Dataset}\vspace{1mm}

The scope of this work is to provide an evaluation framework along with extensive experiments to further study fairness within the legal domain. Following the work of \citet{angwin2016}, \citet{dressel2018}, and \citet{wang-etal-2021-equality}, we provide a diverse benchmark covering multiple tasks, jurisdictions, and protected (examined) attributes.
We conduct experiments based on pre-trained transformer-based language models and compare model performance across four representative group-robust algorithm, i.e., Adversarial Removal \cite{elazar-goldberg-2018-adversarial}, Group DRO \cite{sagawa-etal-2020-dro}, IRM \cite{arjovsky-etal-2020-irm} and REx \cite{krueger-etal-2020-rex}.

We believe that this work can inform and help practitioners to build assisting technology for legal professionals - with respect to the legal framework (jurisdiction) they operate -; technology that does not only rely on performance on majority groups, but also considering minorities and the robustness of the developed models across them. We believe that this is an important application field, where more research should be conducted \cite{tsarapatsanis-aletras-2021-ethical} in order to improve legal services and democratize law, but more importantly highlight (inform the audience on) the various multi-aspect shortcomings seeking a responsible and ethical (fair) deployment of technology.\vspace{2mm}

\noindent\textbf{Credit Attribution / Licensing}\vspace{1mm}

We standardize and put together four datasets: ECtHR \cite{chalkidis-etal-2021-paragraph}, SCOTUS \cite{spaeth2020}, FSCS \cite{niklaus-etal-2021-swiss}, and CAIL \cite{xiao-et-al-2018-cail,wang-etal-2021-equality} that are already publicly available under CC-BY-(NC-)SA-4.0 licenses. We release the compiled version of the dataset under a CC-BY-NC-SA-4.0 license to favor academic research, and forbid to the best of our ability potential commercial dual use.\footnote{\url{https://creativecommons.org/licenses/by-nc-sa/4.0/}} All datasets, except SCOTUS, are publicly available and have been previously published. If datasets or the papers where they were introduced in were not compiled or written by ourselves, we have referenced the original work and encourage FairLex users to do so as well. In fact, we believe that this work should only be referenced, in addition to citing the original work, when jointly experimenting with multiple FairLex datasets and using the FairLex evaluation framework and infrastructure, or use any newly introduced annotations (ECtHR, SCOTUS). Otherwise only the original work should be cited.\vspace{2mm}

\noindent\textbf{Personal Information}\vspace{1mm}

The data is in general partially anonymized in accordance with the applicable national law. The data is considered to be in the public sphere from a privacy perspective. This is a very sensitive matter, as the courts try to keep a balance between transparency (the public's right to know) and privacy (respect for private and family life).
ECtHR cases are partially annonymized by the court. Its data is processed and made public in accordance with the European data protection laws.
SCOTUS cases may also contain personal information and the data is processed and made available by the US Supreme Court, whose proceedings are public.  While this ensures compliance with US law, it is very likely that similarly to the ECtHR any processing could be justified by either implied consent or legitimate interest under European law. In FSCS, the names of the parties have been redacted by the courts according to the official guidelines. CAIL cases are also partially anonymized by the courts according to the courts' policy. Its data is processed and made public in accordance with Chinese Law.
 \vspace{2mm}

\section*{Acknowledgments} This work is fully funded by the Innovation Fund Denmark (IFD)\footnote{\url{https://innovationsfonden.dk/en}} under File No.\ 0175-00011A. We would like to thank the authors of the original datasets for providing access to the original documents, metadata, or confidentially sharing pre-released versions of the datasets.

\bibliography{anthology,custom}
\bibliographystyle{acl_natbib}

\appendix

\section{Discrimination and Fairness in Law}
\label{sec:appendix_discrimination}
The legal notion of \emph{discrimination} has a different scope and semantics in comparison to the notions of \emph{fairness and bias} used in the context of machine learning \cite{gerards2020}, where the aim usually is not to achieve \emph{equal odds}, e.g. that a court shall rule the same decision for both men and woman based on similar facts, or to have 50/50 favourable decisions for both man and woman, but \emph{equal opportunities} \cite{rawls_theory_1971}.

In the context of law, the principle of \emph{equality} and \emph{non-discrimination} is of paramount importance at international, regional and domestic level. Article 2 of the Universal Declaration of Human Rights (UDHR) prohibits discrimination on grounds of race, colour, sex, language, religion, political or other opinion, national or social origin, property, birth or other status, with the latter term having an open-ended meaning. The principle is also reflected in several other United Nations (UN) human rights instruments and in regional legal instruments, including Article 24 American Convention of Human Rights (ACHR), Articles 2 and 3 African Charter on Human and People’s Rights (ACHPR) and Article 14 and Protocol N. 12 of the European Convention on Human Rights (ECHR).

The principle of non-discrimination is included in all international human rights instruments, although only a few explicitly provide a definition of non-discrimination (e.g. Article 1(1) CERD, Article 1 CEDAW, Article 2 CRPD, Article 1(1) ILO). In general, in international human rights law a violation of the principle of non-discrimination occurs when: (a) equal cases are treated differently, (b)  there is no reasonable and objective justification for the difference in treatment, or (c) the means used are not proportional to the aim.
In addition, many international instruments explicitly allow for ‘positive action’, without mandating an obligation on States in that sense. The term ‘positive action’ refers to active measures taken by private institutions or governments that favour members of previously disadvantaged groups with the aim to remedy the effects of past and present discrimination. 
At both regional and domestic level, a great number of countries have implemented non-discrimination law directly in their legislation. The following brief analysis provides an overview of the legal framework applicable in the EU and in the USA, in light of the wide deployment of algorithms and increasing risk of algorithmic discrimination documented in these contexts.

In the context of EU, EU non-discrimination law prohibits both \emph{direct and indirect discrimination}.\footnote{Directives 2000/43/EC of 29 June 2000; 2000/78/EC of 27 November 2000; 2004/113/EC of 13 December 2004; 2006/54/EC of 5 July 2006} \emph{Direct discrimination} occurs when one person is treated ``\emph{less favourably than another is, has been or would be treated in a comparable situation}'' on grounds of sex, racial or ethnic origin, disability, sexual orientation, religion or belief and age in the context of a protected sector (e.g. the workplace and provision of goods and services) \cite{Wachter2021BiasPI}.  Prohibiting direct discrimination allows to provide people with equal access to opportunities (i.e. formal equality). This however does not suffice, nor guarantee to create equality of opportunity (i.e. substantive equality), which can instead be achieved only by accounting for protected attributes and for social and historical realities and by taking positive measures to level the playing field \cite{Fredman2016}.  The notion of \emph{indirect discrimination} is grounded on achieving substantive equality in practice. The Directives define indirect discrimination as situations where an apparently neutral provision, criterion or practice would put persons with a protected characteristic at disadvantage in comparison to other persons, unless ‘that provision, criterion or practice is ``\emph{justified by a legitimate aim and the means of achieving that aim are appropriate and necessary}''. 

Nevertheless, the current EU non-discrimination law framework suffers from limitations, both as regards its personal (i.e. it only protects six characteristics) and material scope (i.e. the prohibition on discrimination is limited only to certain fields) \cite{gerards2020}.  These limitations pose problems in connection to algorithmic discrimination. For example, as algorithmic bias often creates seemingly neutral distinctions which however often correlate to a protected group (i.e. proxy discrimination), the limited list of protected grounds renders difficult to tackle the effects of algorithmic bias through the concept of direct discrimination \cite{prince2019}.  Indirect discrimination can help address those cases. but its application in this context poses several challenges. 

In April 2021 the European Commission presented a proposal for a Regulation laying down harmonized rules on artificial intelligence (AI Act / AIA).\footnote{Regulation Proposal 2021/206} The proposal aims at avoiding ``\emph{significant risks to the health and safety or fundamental rights of persons}''  and would, once adopted, complement the currently applicable legal framework for tackling algorithmic discrimination, thereby overcoming some of its existing limitations. The envisaged implementation of the proposed AI Act highlights the importance that the legislator poses in preventing and mitigating discrimination and biases arising from the development and use of AI systems in several areas of application, including in the legal sector \cite{schwemer2021}.  AI systems used for the administration of justice and democratic processes are proposed to be deemed high-risk in order ``\emph{to address the risks of potential biases, errors, and opacity}'' (recital 40 AIA). The consequence is that such systems would be subject to a variety of design and development requirements, e.g. related to the training, validation and testing data sets which would have to be examined \emph{inter alia} in relation to possible biases (art. 10(2) lit. f AIA) or related to human oversight of such AI system with a view to remain aware of automation bias (art. 14(4) lit. b AIA).

In the US the jurisprudence relies on the doctrines of \emph{disparate treatment} and \emph{disparate impact} provided for in Title VII of the 1964 Civil Rights Act.\footnote{Civil Rights Act of 1964, 42 U.S.C § 2000e-2}  A prohibition on disparate treatment  is included also in the Equal Protection Clause of the Constitution\footnote{Cf. Vasquez v. Hillery, 474 US 254 (1986).} and in civil rights laws. The prohibition refers to intentional discrimination and occurs when \emph{individuals are treated in a different manner on the basis of protected class attributes}, such as race, colour, national origin, sex, age and religion.\footnote{Civil Rights Act of 1964, 42 U.S.C § 2000e-2} The prohibition on disparate impact is instead provided for only in civil rights statutes and, in brief, it establishes that if some practice or activity has a disproportionate adverse effect on protected groups, the defendant must prove that such a practice has an adequate justification.\footnote{See the defining decision, Griggs v. Duke Power Co., 401 U.S. 424 (1971).}  Also in the US, recent literature has highlighted the challenges that the current legal framework faces when tackling algorithmic discrimination, in particular as far as liability and the burden of proof are concerned \cite{kleinberg-etal-2019,xiang-2021}.  

Beyond the boundaries of EU and US law, a great number of countries explicitly prohibit discrimination in their laws on the basis of nationality, race, ethnicity and religion. Other countries extend the prohibition only in relation to race and religion instead. In many countries, there is not yet any specific or dedicated law against non-discrimination, such as in China, India, Indonesia, Japan, Korea and Saudi Arabia. This does not imply by any means that there are not potentially separate pieces of legislation that enforce non-discrimination for some class attributes. 

\begin{figure*}
    \centering
    \resizebox{\textwidth}{!}{
    \includegraphics{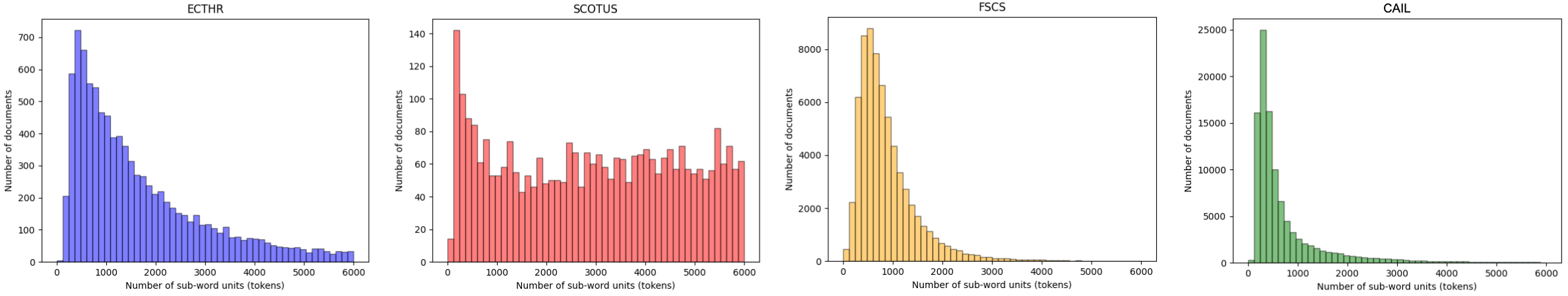}
    }
    \vspace{-3mm}
    \caption{Distribution of sequence (document) length across FairLex datasets (ECtHR, SCOTUS, FSCS, CAIL).}
    \label{fig:datasets}
    \vspace{-4mm}
\end{figure*}

\section{Attribute Extraction and Grouping}
\label{sec:attributes}

In this section, we provide finer details on attribute extraction and grouping.

\paragraph{ECtHR}
We extracted the defendant states from the HUDOC\footnote{The ECtHR online database (\url{https://hudoc.echr.coe.int/})} case metadata, namely \emph{Respondent State(s)}. We group the defendant states mainly relying on their classification  by the EuroVoc thesaurus\footnote{\url{https://op.europa.eu/en/web/eu-vocabularies}}. The grouping mainly reflects the high disproportion of violations between mainly Eastern European countries, and in a second degree Central European, and the rest (Western European, Nordic, Mediterranean states). Applicant's birth year is extracted from the case facts, if available, e.g., \emph{``The first applicant, Mr X, was born in \textbf{1967}.''}, using Regular Expressions (RegEx). Then, we compute the age by subtract birth year from the judgment date, extracted from the HUDOC case metadata, as well. The age grouping does not follow any pattern and aims to cluster applicants in discrete groups that have statistical support. Finally, we extract gender from case facts, if possible, based on pronouns (e.g., `he', `she', `his', `her'), and other gender words (e.g., `mr', `mrs', `husband', `wife') in context such as \emph{``The applicant's \textbf{husband} [...]''}, or \emph{```The applicant, \textbf{Mr} A, [...]''}. We acknowledge that non-binary gender identities exist, but non-binary gendered applicants cannot be identified automatically. 

In many cases, the birth year or gender was not identifiable in the facts. Furthermore, many cases involve multiple applicants. In such cases, we mark the respected attributes as unknown, and hold a different group for unidentified instances. These data points are used in experiments, but we do not report results for such groups.

\paragraph{SCOTUS}

Both attributes rely on metadata provided by the Supreme Court DataBase  (SCDB)\footnote{\url{http://scdb.wustl.edu/}}. In case of the \emph{direction of the decision}, i.e., whether the decision is considered liberal, or conservative, we use the original variable (\emph{Decision Direction}).\footnote{\url{http://scdb.wustl.edu/documentation.php?var=decisionDirection}}. In case of the \emph{type of respondent}, we manually categorize (cluster) all available, 311 in total, values for the \emph{Respondent} variable\footnote{\url{http://scdb.wustl.edu/documentation.php?var=respondent}} in five abstract categories (person, public entity, organization, facility and other).

\paragraph{FSCS}

All attributes are already available as part of the original dataset of \citet{niklaus-etal-2021-swiss}. Groups represent individual values. Information was extracted from courts' metadata.

\paragraph{CAIL}

All attributes are already available as part of the original dataset of \citet{wang-etal-2021-equality}. Groups represent individual values. Information was extracted from courts' metadata.

\section{Train and Evaluation Details} 

We fine-tune all pre-trained transformer-based language models using the AdamW \cite{loshchilov2018decoupled} optimizer with a learning rate of 3e-5. We use a batch size of 16 and train models for up to 20 epochs using early stopping on validation performance.\footnote{We train all models in a mixed-precision (fp16) setting to use the maximum available batch size.} Across datasets and attributes, we run five repetitions with different random seeds and report averaged scores.

\section{Release of Language Models}

We release four domain-specific BERT models (Table~\ref{tab:models} with continued pre-training on the corpora of the examined datasets.\footnote{\url{https://huggingface.co/coastalcph}} We train mini-sized BERT models with 6 Transformer blocks, 384 hidden units, and 12 attention heads. We warm-start all models from the public MiniLMv2 models checkpoints \cite{wang-etal-2021-minilmv2} using the distilled version of RoBERTa~\cite{liu-etal-2019-roberta} for the English datasets (ECtHR, SCOTUS) and the one distilled from XLM-R~\cite{conneau-etal-2020-unsupervised} for the rest (trilingual FSCS, and Chinese CAIL). We pre-train each models in the training subset of each FairLex dataset with sequences of 128 sub-words for 10 epochs using the AdamW \cite{loshchilov2018decoupled} optimizer with a maximum learning rate of 1e-4 and 10\% warm-up ratio.

\begin{table}[]
     \centering
    \resizebox{\columnwidth}{!}{
    \begin{tabular}{l|l|l}
     Model name  & Domain & Languages \\
     \hline
 `coastalcph/fairlex-ecthr-minlm`  & ECtHR            & `en`               \\
 `coastalcph/fairlex-scotus-minlm` & SCOTUS           & `en`               \\
 `coastalcph/fairlex-fscs-minlm`   & FSCS             & `de`, `fr`, `it` \\
 `coastalcph/fairlex-cail-minlm`   & CAIL             & `zh`               \\

    \end{tabular}
    }
    \caption{Domain-specific pre-trained language models specifications.}
    \label{tab:models}
    \vspace{-4mm}
\end{table}

\section{Statistics}
\subsection{Distribution of Document Length}
\label{sec:appendix_b}
In Figure~\ref{fig:datasets} we report the distribution of sequence (document) length across FairLex datasets (ECtHR, SCOTUS, FSCS). We observe that the documents are extremely long (3,000-6,000+ words) across datasets. Hence, we deploy hierarchical models (Section~\ref{sec:models}) that are able to encode large parts of the documents.

\subsection{Group Distribution by Attribute}
\label{sec:group_dist}
In Tables \ref{tab:ecthr_group_dist} and \ref{tab:scotus_group_dist} we report the group distribution per examined attribute under consideration. In some cases, the extraction of the specific attribute, e.g., gender or age in ECtHR, was not possible, i.e., the applied rules would no suffice, possibly because the information is intentionally missing. During training, the groups of unidentified samples is included, but we report test scores excluding those, i.e., $\overline{\mathrm{mF1}}$ and $GD$ do not take into account the F1 of these groups.

\section{Label Distribution KL Divergences}
\label{sec:appendix_divergence}
In Tables~\ref{tab:js_ecthr}, \ref{tab:js_scotus}, \ref{tab:js_fscs}, and \ref{tab:js_spc}, we report the Jensen-Shannon divergences between train-test, train-dev and test-test distribution of labels separately for each protrected attribute values and for each dataset in our framework. 

\begin{table*}[t]
    \centering
    \begin{tabular}{rrrr|rrr|rr}
    \toprule
    \multicolumn{9}{c}{\textsc{ECtHR}} \\
    \midrule
    \multicolumn{4}{c|}{\textit{Applicant Age}} & \multicolumn{3}{c|}{\textit{Applicant Gender}} & \multicolumn{2}{c}{\textit{Defendant State}} \\
    N/A & $\leq35$ & $\leq 65$ & $>65$ & N/A & Male & Female & E.C. & West \\
    \midrule
    2,794 & 839 & 4,246 & 1,121 & 3,306 & 4,407 & 1,287 & 7,224 & 1,776 \\
    \bottomrule
    \end{tabular}
    \caption{Group distribution in training set for each attribute of ECtHR dataset. `N/A' (Not Answered) refers to samples, where the respected attribute could not be extracted.}
    \label{tab:ecthr_group_dist}
\end{table*}

\begin{table*}[t]
    \centering
    \begin{tabular}{rrrrr|rr} 
    \midrule
    \multicolumn{7}{c}{\textsc{SCOTUS}}\\
    \midrule
    \multicolumn{5}{c|}{\textit{Defendant}} & \multicolumn{2}{c}{\textit{Direction}} \\
    Other & Facility & Organization & Person & Public Entity & Conservative & Liberal \\
    \midrule
    957 & 140 & 741 & 1847 & 2796 & 3146 & 3335\\
    \bottomrule
    \end{tabular}
    \caption{Group distribution in training set for each attribute of SCOTUS dataset.}
    \label{tab:scotus_group_dist}
\end{table*}

\begin{table*}[]
    \centering
    \begin{tabular}{l|ccc|cc|cc}
    \toprule
    & \multicolumn{3}{c|}{\textit{Applicant Age}} & \multicolumn{2}{c|}{\textit{Applicant Gender}} & \multicolumn{2}{c}{\textit{Defendant State}} \\
    & $\leq35$ & $\leq 65$ & $>65$ & Male & Female & East & West \\
    \midrule
    Train-Test & 0.19 & 0.18 & 0.32 & 0.17 & 0.26 & 0.17 & 0.28 \\
    Train-Dev &  0.18 & 0.19 & 0.22 & 0.17 & 0.22 & 0.18 & 0.17  \\
    Dev-Test & 0.20 & 0.08 & 0.19 & 0.09 & 0.10 & 0.09 & 0.16 \\
    \bottomrule
    \end{tabular}
    \caption{Jensen-Shannon Divergence of label distribution between training, test and development sets of ECtHR by protected attribute values. The lower the values the more similar the distributions.}
    \label{tab:js_ecthr}
\end{table*}

\begin{table*}[t]
    \centering
    \begin{tabular}{l|ccccc|cc}
    \toprule
    & \multicolumn{5}{c|}{\textit{Defendant}} & \multicolumn{2}{c}{\textit{Direction}} \\
    & Facility & Organization & Other & Person & Pub. Entity & Conservative & Liberal \\
    \midrule
    Train-Test & 0.26 & 0.11 & 0.09 & 0.05 & 0.07 & 0.05 & 0.04\\
    Train-Dev & 0.28 & 0.11 & 0.11& 0.07 & 0.03 &0.06 & 0.05 \\
    Dev-Test & 0.22 & 0.17 & 0.13 & 0.10 & 0.07 & 0.09 & 0.07 \\
    \bottomrule
    \end{tabular}
    \caption{Jensen-Shannon Divergence of label distribution between training, test and development set in Scotus by protected attribute values. The lower the values the more similar the distributions.}
    \label{tab:js_scotus}
\end{table*}

\begin{table*}[t]
    \centering
    \begin{tabular}{ll|ccc}
    \toprule
    & & Train-Test & Train-Dev & Dev-Test \\
    \midrule
    
    \multirow{3}{*}{\textit{Language}} & DE & 0.0336 & 0.0275 & 0.0061\\
    & FR & 0.0517 & 0.0301 & 0.0216\\
    & IT & 0.0145 & 0.0405& 0.0261\\
    \midrule
    \multirow{6}{*}{\textit{Legal Area}} & Other &0.1000 & --- & ---\\
    & Public Law & 0.0007 & 0.0090 & 0.0083\\
    & Penal Law & 0.0018 &0.0118 & 0.0136\\
    & Civil Law & 0.0248 & 0.0046& 0.0202\\
    & Social Law & 0.0624 & 0.0570 & 0.0054\\
    \midrule
    \multirow{8}{*}{\textit{Region}} & Région lémanique & 0.0447 & 0.0259 & 0.0188\\
    & Zürich &0.0447 & 0.0345 & 0.0028\\
    & Espace Mittelland & 0.0765 &0.0435 & 0.0331\\
    & NW Switzerland & 0.0280& 0.0127 & 0.0407\\
    & E Switzerland & 0.0197 & 0.0394 & 0.0198\\
    & C Switzerland & 0.0267 & 0.0304 & 0.0036\\
    & Ticino & 0.0023 & 0.0284& 0.0307 \\
    & Federation & 0.0018& 0.0385& 0.0404\\
    \bottomrule
    \end{tabular}
    \caption{Jensen-Shannon Divergence of label distribution between training, test and development set in FSCS by protected attribute values. The lower the values the more similar the distributions.}
    \label{tab:js_fscs}
\end{table*}

\begin{table*}[t]
    \centering
    \begin{tabular}{l|ccccccc|cc}
    \toprule
    & \multicolumn{7}{c|}{\textit{Region}} & \multicolumn{2}{c}{\textit{Gender}} \\
    & Beijing & Liaoning & Hunan & Guangdong & Sichuan & Guangxi & Zhejiang & Male & Female \\
    \midrule
    Train-Test &0.0516 & 0.0458 & 0.0495 &0.0524 & 0.0559 & 0.0696 & 0.0687 & 0.0345 & 0.0766 \\
    Train-Dev &0.0239 & 0.0270 & 0.0406 & 0.0584 & 0.0484 & 0.0426 & 0.0338 & 0.0164 &0.0318 \\
    Dev-Test & 0.0469 & 0.0296& 0.0799 & 0.0431 & 0.0554 & 0.0496 & 0.0633 & 0.0307 & 0.0986 \\
\bottomrule
    \end{tabular}
    \caption{Jensen-Shannon Divergence of label distribution between training, test and development set in SPC by protected attribute values. The lower the values the more similar the distributions.}
    \label{tab:js_spc}
\end{table*}

\end{document}